\documentclass[conference, a4paper]{IEEEtran}
\usepackage[a4paper, top=1.9cm, bottom=4.4cm, left=1.35cm, right=1.35cm]{geometry}
\IEEEoverridecommandlockouts
\usepackage[utf8]{inputenc}
\usepackage{CJK}
\usepackage{cite}
\usepackage{amsmath,amssymb,amsfonts}
\usepackage{algorithmic}
\usepackage{graphicx}
\usepackage{textcomp}
\usepackage{xcolor}
\usepackage{marvosym}
\def\BibTeX{{\rm B\kern-.05em{\sc i\kern-.025em b}\kern-.08em
    T\kern-.1667em\lower.7ex\hbox{E}\kern-.125emX}}
\begin{document}
\title{Multiple-Debias: A Full-process Debiasing Method for Multilingual Pre-trained Language Models}

\author{
\IEEEauthorblockN{Haoyu Liang$^{1}$, Peijian Zeng$^{1}$, Wentao Huang$^{1}$, Aimin Yang$^{1,}$\textsuperscript{\Letter}, Dong Zhou$^{2,}$\textsuperscript{\Letter}}

\IEEEauthorblockA{
\textit{$^{1}$ School of Computer Science and Technology}, \textit{Guangdong University of Technology}, Guangzhou, China \\
\textit{$^{2}$ School of Information Science and Technology}, \textit{Guangdong University of Foreign Studies}, Guangzhou, China \\
amyang@gdut.edu.cn, dongzhou@gdufs.edu.cn}
\thanks{\textsuperscript{\Letter} Aimin Yang and Dong Zhou are the co-corresponding authors.}
}

\maketitle

\begin{abstract}
Multilingual Pre-trained Language Models (MPLMs) have become essential tools for natural language processing. However, they often exhibit biases related to sensitive attributes such as gender, race, and religion. In this paper, we introduce a comprehensive multilingual debiasing method named Multiple-Debias to address these issues across multiple languages. By incorporating multilingual counterfactual data augmentation and multilingual Self-Debias across both pre-processing and post-processing stages, alongside parameter-efficient fine-tuning, we significantly reduced biases in MPLMs across three sensitive attributes in four languages. We also extended CrowS-Pairs to German, Spanish, Chinese, and Japanese, validating our full-process multilingual debiasing method for gender, racial, and religious bias. Our experiments show that (i) multilingual debiasing methods surpass monolingual approaches in effectively mitigating biases, and (ii) integrating debiasing information from different languages notably improves the fairness of MPLMs. 
\end{abstract}


\section{Introduction}
In recent years, Pre-trained Language Models (PLMs) like BERT \cite{DBLP:conf/naacl/DevlinCLT19} and GPT-3 \cite{DBLP:conf/nips/BrownMRSKDNSSAA20} have demonstrated outstanding performance in Natural Language Processing (NLP) tasks, from text classification to machine translation. Advanced Large Language Models (LLMs), including GPT-4 \cite{DBLP:conf/nips/BrownMRSKDNSSAA20} and LLaMA \cite{touvron2023llama}, have further achieved state-of-the-art results across various NLP applications. These models are trained unsupervised on large-scale data, enabling them to capture rich semantic and syntactic information, greatly enhancing their generalization abilities.

However, the strengths of these PLMs come with notable drawbacks. Biases present in the training data can lead to biased inferences, resulting in unfair model outputs \cite{DBLP:conf/acl/HutchinsonPDWZD20}. These models rely heavily on large datasets often sourced from internet content. As a result, they inevitably inherit social biases, including race, gender, or language, which can be retained or even amplified in downstream tasks. To address this challenge, researchers have developed a range of debiasing techniques aimed at reducing unfairness in PLMs. These techniques are categorized into pre-processing, in-processing, and post-processing methods \cite{pessach2022review}. Nevertheless, most existing studies focus on single languages or specific attributes like gender or race, with limited attention to multi-language or multi-attribute settings \cite{xu2024survey}. 

In response to this challenge, this paper introduces a new full-process debiasing method, named Multiple-Debias (MD), to mitigate biases in PLMs across multiple languages and sensitive attributes. Our approach first extends Counterfactual Data Augmentation (CDA) \cite{DBLP:conf/naacl/ZhaoWYOC18} to support four languages that represent both Eastern and Western cultures: German, Spanish, Chinese, and Japanese. It also addresses three sensitive attributes, including gender, race, and religion. We subsequently utilize this expanded multilingual and multi-attribute CDA corpus alongside full fine-tuning and three Parameter-Efficient Fine-Tuning (PEFT): adapter-tuning \cite{houlsby2019parameter}, prefix-tuning \cite{DBLP:conf/acl/LiL20}, and prompt-tuning \cite{DBLP:conf/emnlp/LesterAC21}. By using this approach, we fine-tune two Multilingual Pre-trained Language Models (MPLMs), mBERT \cite{DBLP:conf/naacl/DevlinCLT19} and XLM-R \cite{DBLP:conf/acl/ConneauKGCWGGOZ20}. Finally, we incorporate an extended version of Self-Debias (SD)  \cite{schick2021self} for multilingual settings to complete a three-stage debiasing process. We also adapted the CrowS-Pairs \cite{DBLP:conf/emnlp/NangiaVBB20} bias evaluation method into the four target languages using machine translation. Experimental results demonstrate that our approach achieves substantial debiasing effects across multiple languages and sensitive attributes.

Our main contributions are as follows:

\begin{itemize}
% \item We present a comprehensive debiasing framework, termed Multiple-Debias, which combines Multilingual Counterfactual Data Augmentation and Multilingual Self-Debias techniques with Parameter-Efficient Fine-Tuning. This framework addresses multiple sensitive attributes, including gender, race, and religion, across various languages, offering a scalable solution for effectively reducing biases in MPLMs.

\item We propose Multiple-Debias, a scalable framework that combines multilingual counterfactual data augmentation, multilingual Self-Debias, and parameter-efficient fine-tuning to reduce gender, race, and religion biases in MPLMs across languages.

\item To evaluate the effectiveness of our approach, we introduce the multilingual CrowS-Pairs metric, extending the CrowS-Pairs dataset to German, Spanish, Chinese, and Japanese. This expanded evaluation dataset enables more robust assessment across diverse languages and cultural contexts, highlighting the adaptability of our framework to multilingual settings.
\item Our findings suggest that integrating fairness data across languages significantly improves model performance in mitigating multilingual biases. This insight underscores the importance of multilingual debiasing information in enhancing the fairness of MPLMs.
\end{itemize}

% \addtolength{\topmargin}{0.054cm}
\section{Multiple-Debias}
\begin{figure*}
        % \vspace{0.04 in} 
        \centering
	\includegraphics[scale=.54]{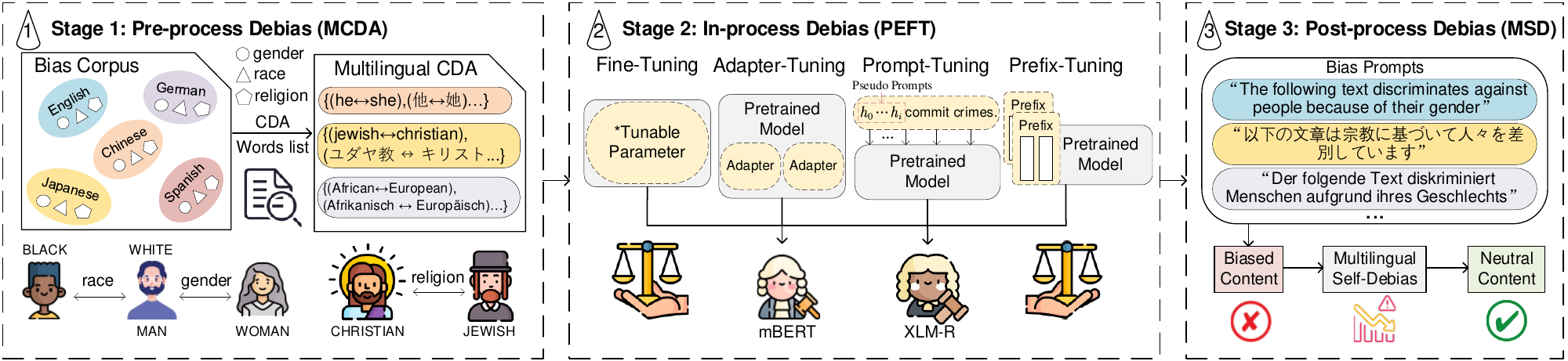}
        \caption{The framework of Multiple-Debias.}
	\label{FIG:1}
\end{figure*}

Most existing research on debiasing has centered on single-language models, with few studies addressing the complexities of multilingual debiasing. Prior approaches to assessing and mitigating bias in multilingual models has largely focused on reducing gender bias in cross-lingual word embeddings \cite{DBLP:conf/acl/ZhaoMHCA20}. However, these approaches did not take into account how word meanings can change in different contexts, limiting its reliability for application in PLMs. Some recent works have expanded CDA \cite{DBLP:conf/naacl/ZhaoWYOC18} and Self-Debias \cite{schick2021self} methods to include languages beyond English, enhancing their practicality in multilingual contexts. For instance, Vashishtha et al. \cite{DBLP:conf/acl/VashishthaAS23} adapted these methods to six Indian languages, while Reusens et al. \cite{DBLP:conf/emnlp/ReusensBMWB23} demonstrated that many monolingual debiasing techniques can transfer effectively across languages. Even so, bias can arise at various stages of language model processing, including training data, input representation, model structure, and research design. Therefore, addressing bias at just one stage is often inadequate \cite{DBLP:conf/eacl/RameshSC23}. To tackle this complexity, we propose a novel, comprehensive approach to debiasing across multiple languages and sensitive attributes which we call Multiple-Debias (MD). As shown in Fig. \ref{FIG:1}, MD consists of three key modules:

\textbf{1. Pre-Processing Debiasing}: We begin by translating sensitive terms related to gender, race, and religion from English into four other languages (Chinese, German, Spanish, and Japanese) to represent a range of Eastern and Western cultures. Using these terms, we construct a multilingual corpus drawn from Wikipedia in five languages as the basis for our Multilingual Counterfactual Data Augmentation (MCDA). Applying the CDA algorithm, we then create a training corpus covering multiple languages and attributes.

\textbf{2. In-Process Debiasing}: Next, we apply MCDA to train MPLMs, specifically mBERT and XLM-R, using three Parameter-Efficient Fine-Tuning (PEFT) techniques: adapter-tuning, prompt-tuning, and prefix-tuning. This step aims to mitigate bias during model training with multilingual and multi-attribute awareness.

\textbf{3. Post-Processing Debiasing}: Finally, we implement Self-Debias (SD) templates tailored to each of the five languages to complete the debiasing process. This Multilingual Self-Debias (MSD) technique allows for an additional layer of bias reduction after generation.

Our experimental results show that Multiple-Debias achieves effective debiasing in multilingual languages and sensitive attributes. Additionally, we observe that this multilingual optimization yields improvements beyond those seen in single-language debiasing methods.

% \addtolength{\topmargin}{0.02 in}

\subsection{Multilingual Counterfactual Data Augmentation}
We introduce Multilingual Counterfactual Data Augmentation (MCDA) as the pre-processing debiasing module in the Multiple-Debias framework, extending the concept of CDA by incorporating multilingual text data. CDA, first proposed by Zhao et al. \cite{DBLP:conf/naacl/ZhaoWYOC18}, is a data-driven pre-processing method used to reduce bias related to sensitive attributes like gender, race, and religion. Its core idea is to balance the dataset by swapping words associated with biased attributes (e.g., \textit{``he"} and \textit{``she"}) to create a more balanced corpus. For instance, to reduce gender bias, the sentence “The teacher entered the classroom, and he began to teach” could be augmented as “The teacher entered the classroom, and she began to teach.” This generates a balanced dataset for further model training, effectively reducing bias. 

We integrated multilingual information into CDA with the aim of enhancing its performance in reducing biases across multiple languages. Unlike Vashishtha et al. \cite{DBLP:conf/acl/VashishthaAS23}, who applied debiasing first and then translated parallel corpus, we directly use Wikipedia dumps in five languages (English, Chinese, Japanese, German, and Spanish) as our original corpus. This allows us to preserve the original cultural context of each language, enhancing MCDA's debiasing effectiveness across languages. To construct a multilingual and multi-attribute CDA corpus, we first used the DeepL translator to translate an English list of sensitive terms covering gender, race, and religion into four additional languages. For quality assurance, we manually reviewed the Chinese and German translations, confirming their accuracy and reliability. We then applied text augmentation to these multilingual corpus by symmetrically replacing certain terms, such as \textit{“gender: \{(he$\leftrightarrow$she), \begin{CJK}{UTF8}{gbsn}(他$\leftrightarrow$她)\}", \end{CJK}}\textit{``religion: \{(Jewish$\leftrightarrow$Christian), 
 \begin{CJK}{UTF8}{gbsn}(ユダヤ教$\leftrightarrow$キリスト教)\}", \end{CJK}}and \textit{``race: {(African$\leftrightarrow$European), (Afrikanisch$\leftrightarrow$Europ\"aisch)}."} By systematically replacing sensitive terms throughout the entire multilingual corpus, we generated a new, balanced dataset for further training.

\subsection{Parameter-Efficient Fine-Tuning}
Traditional monolingual CDA methods typically rely on full fine-tuning, which is both time-consuming and resource-intensive. As language models grow in size, they become more prone to biased behavior, and debiasing them becomes increasingly challenging due to the higher computational costs involved. Xie et al. \cite{DBLP:conf/acl/XieL23} demonstrated that Parameter-Efficient Fine-Tuning (PEFT) can effectively address both the issue of bias and the associated computational burden. PEFT can modify only specific parameters, such as those related to harmful biased knowledge encoded during pre-training, while preserving model performance.

For this reason, we integrate PEFT as the in-processing debiasing module within MCDA during the debiasing process. This allows the model to effectively encode fairness and eliminate harmful biases introduced during pre-training. To implement this, we apply three PEFT techniques: adapter tuning \cite{houlsby2019parameter}, prefix tuning \cite{DBLP:conf/acl/LiL20}, and prompt tuning \cite{DBLP:conf/emnlp/LesterAC21}, to train MPLMs with the MCDA corpus. We then compare these results with traditional full fine-tuning to assess their effectiveness. Using these methods, we fine-tune mBERT and XLM-R to remove biases embedded in their pre-trained knowledge. This approach reduces the time and computational resources needed for fine-tuning while enhancing model fairness with minimal impact on performance.

\subsection{Multilingual-Self-Debias}
Schick et al. \cite{schick2021self} proposed the Self-Debias method, which leverages the inherent capacity of PLMs to recognize their own biases. This technique works by first prompting the model to generate biased content using specific prompts, such as \textit{``The following text discriminates against people because of their gender."} The model then revises this biased output by reducing the prominence of biased terms, effectively suppressing their occurrence in the final text. This process begins by prompting the model to produce biased statements because models generally struggle to respond accurately to negative prompts, like \textit{``The following text contains no bias."} By first generating biased text and then suppressing these biases, the Self-Debias method more effectively mitigates bias in the output.

Multilingual-Self-Debias (MSD) extends this concept by translating the English prompts, such as \textit{``The following text discriminates against people because of their gender/race or color/religion"} into different languages. For example, it becomes\begin{CJK}{UTF8}{gbsn}“以下文字因性别歧视人”\end{CJK}in Chinese, \textit{``Der folgende Text diskriminiert Menschen aufgrund ihrer Rasse/Farbe"} in German, or\begin{CJK}{UTF8}{gbsn}“以下の文章は宗教に基づいて人々を差別しています”\end{CJK}in Japanese. These translated prompts guide the model to better recognize and mitigate biases in the text across different languages. To apply this approach, we created customized Self-Debias templates in four languages, targeting three different sensitive attributes. This multilingual approach serves as the post-processing debiasing module in our Multiple-Debias framework, ensuring that the final generated content is effectively debiased.

\section{Multilingual Fairness Evaluation}
Monolingual evaluation metrics are relatively well-established, with a range of standardized tests and benchmarking tools available. For example, Crowdsourced Stereotype Pairs (CrowS-Pairs) \cite{DBLP:conf/emnlp/NangiaVBB20} and StereoSet \cite{DBLP:conf/acl/NadeemBR20} are both commonly used to detect and quantify biases within language models. However, as research on multilingual debiasing progresses, these monolingual evaluation methods are proving inadequate for addressing the complexities of bias in multilingual contexts. To address this challenge, researchers have begun to develop multilingual bias evaluation tools, such as Multilingual DisCo \cite{DBLP:conf/acl/VashishthaAS23} and Multilingual Bias Evaluation (MBE) \cite{DBLP:conf/naacl/KanekoIBO22}. However, since these tools do not fully meet our needs for evaluating debiasing across multiple languages and sensitive attributes, we proposed Multilingual CrowS-Pairs, a new framework designed specifically for this purpose.

\subsection{Multilingual CrowS-Pairs}
We developed Multilingual CrowS-Pairs as an extension of CrowS-Pairs \cite{DBLP:conf/emnlp/NangiaVBB20}, a benchmark that tests for bias by comparing sentence pairs. Each pair contains two similar sentences with only slight word differences: one reflecting a stereotype about a disadvantaged group (e.g., \textit{``People who live in trailers are all alcoholics"}), and the other offering a counter-stereotype (e.g., \textit{``People who live in mansions are all alcoholics"}). CrowS-Pairs measures the likelihood that the model favors the stereotypical sentence over the counter-stereotypical one by comparing the masked token probabilities of key words in both sentences. If the model consistently assigns higher probabilities to stereotypical sentences, it indicates a bias.

Although CrowS-Pairs is a well-established tool for fairness evaluation, it only supports English, limiting its use for assessing fairness in multilingual settings. Reusens et al. \cite{DBLP:conf/emnlp/ReusensBMWB23} expanded CrowS-Pairs further by translating samples from the original English dataset into French, German, and Dutch to assess the fairness of MPLMs. However, their version only covers Western languages, excluding languages such as Chinese and Japanese from the evaluation. To fill this gap, we translated the dataset into four additional languages, including Chinese and Japanese, using the DeepL translator. This allowed us to evaluate our debiasing methods in both Eastern and Western multilingual contexts.

\subsection{Multilingual Bias Evaluation}
We also adopted the Multilingual Bias Evaluation (MBE) \cite{DBLP:conf/naacl/KanekoIBO22} benchmark to validate our multilingual debiasing method. MBE assesses gender bias in MPLMs using parallel corpus and a predefined list of gendered English terms, eliminating the need for annotated data in the target language. The evaluation process in MBE begins by extracting sentences containing gendered terms (e.g., \textit{``father"} and \textit{``mother"}) from parallel corpus between English and the target language. The model under evaluation is then used to compute the likelihood of gendered words in these sentences. MBE pairs sentences containing female terms with those containing male terms and compares their likelihoods. If the likelihood of a male-gendered term is higher than its female counterpart, the pair is marked as biased (scored as 1); otherwise, it is marked unbiased (scored as 0). To calculate the MBE score, these binary results are weighted by the similarity of each gendered sentence pair, and the weighted bias scores are normalized. An MBE score closer to 50\% indicates greater gender balance.

However, MBE is limited to evaluating only gender bias, neglecting other important demographic attributes such as race and religion. Additionally, MBE does not provide an evaluation method for English. For these reasons, we only validated the results of the multilingual debiasing method on the MBE evaluation metrics in four additional languages.

\section{Experiments and Results}
\subsection{Experimental Setup}
In our experiments, we debiased mBERT and XLM-R, using their base checkpoints from Hugging Face Transformers \cite{wolf-etal-2020-transformers}. To ensure balanced multilingual representation, we downsampled Wikipedia data across five languages to an average of 26.8 million tokens per language, combining them into a single corpus. Training was conducted on an NVIDIA GeForce RTX 3090 GPU with 24 GB of memory, following the setup by Zhongbin Xie et al. \cite{DBLP:conf/acl/XieL23}. Each model was trained for two epochs with a random seed of 42, and all other training hyperparameters were kept at their default settings. 

Using the Multilingual CrowS-Pairs and MBE \cite{DBLP:conf/naacl/KanekoIBO22} evaluation metrics, we compared the debiasing performance of the monolingual methods CDA \cite{DBLP:conf/naacl/ZhaoWYOC18} and SD \cite{schick2021self} with our multilingual methods MCDA, MSD, and MD on mBERT \cite{DBLP:conf/naacl/DevlinCLT19} and XLM-R \cite{DBLP:conf/acl/ConneauKGCWGGOZ20}. We also examined the impact of Parameter-Efficient Fine-Tuning (PEFT) versus full fine-tuning on the effectiveness of Multiple-Debias.

We report $|50 - \text{Metric Score}|$ as bias score in our result. A bias score closer to 0 indicates less biased model representations. The best score of all the debiasing methods for each language is marked in bold. The final column shows the average score across all four languages for each method.

\subsection{Results of Mitigating Gender Bias}
\textbf{Multilingual debiasing methods effectively reduced gender bias, outperforming monolingual methods.} Using the Multilingual CrowS-Pairs evaluation metrics, we compared different debiasing methods on gender-sensitive attributes, with results shown in Table \ref{table1}. Both MCDA and MSD outperformed their monolingual counterparts, CDA and Self-Debias, on the mBERT model. For mBERT, monolingual CDA struggled, particularly in Spanish, while MD showed consistent improvements across multiple languages. A similar pattern appeared with XLM-R, where adapter-tuned MD achieved the best debiasing results. These findings highlight that multilingual training data and tailored prompts can enhance CDA and SD effectiveness across languages.

\begin{table}[]
\centering
\caption{Evaluation of Gender Bias on Multilingual CrowS-Pairs}
\resizebox{\linewidth}{!}{
\begin{tabular}{lccccc}
\hline
Model                 & DE  & ZH  & ES  & JA  & Avg. ($\downarrow$) \\ \hline
mBERT                 & 8.48            & 2.50            & 4.26            & 3.58            & 4.71              \\ \hline
\hspace{3pt}+CDA                  & 12.73           & 0.83            & 22.81           & 3.49            & 9.97              \\
\hspace{3pt}+SD                   & 8.46            & 5.83            & 2.92            & 6.04            & 5.81              \\ \hline
\hspace{3pt}+MCDA                 & 10.61           & 2.71            & 3.78            & 2.31            & 4.85              \\
\hspace{3pt}+MSD                  & \textbf{0.83}   & 3.33            & 2.98            & 3.43            & 2.64              \\
\hspace{3pt}+MD w/ Full Fine-Tune & 5.06            & 4.17            & 0.92            & \textbf{0.50}   & 2.66              \\
\hspace{3pt}+MD w/ Adapter Tune   & 11.88           & \textbf{0.00}   & \textbf{0.77}   & 2.73            & 3.85              \\
\hspace{3pt}+MD w/ Prompt Tune    & 3.38            & 2.50            & 2.41            & 0.69            & \textbf{2.24}     \\
\hspace{3pt}+MD w/ Prefix Tune    & 2.59            & 1.67            & \textbf{0.77}   & 7.23            & 3.06              \\ \hline \hline
XLM-R                  & 7.35            & 8.33            & 8.60            & 3.56            & 6.96              \\ \hline
\hspace{3pt}+CDA                  & \textbf{1.37}   & 4.17            & \textbf{0.26}   & 6.98            & 3.20              \\
\hspace{3pt}+SD                   & 4.74            & 10.00           & 0.60            & \textbf{0.07}   & 3.85              \\ \hline
\hspace{3pt}+MCDA          & 6.31            & 3.96            & 3.34            & 5.33            & 4.73              \\
\hspace{3pt}+MSD           & 2.11            & 11.67           & 1.86            & 2.71            & 4.59              \\
\hspace{3pt}+MD w/ Full Fine-Tune     & 3.10            & 1.67            & 5.03            & 0.14            & 2.49              \\
\hspace{3pt}+MD w/ Adapter Tune     & 4.74            & \textbf{0.00}   & 2.99            & 0.36            & \textbf{2.02}     \\
\hspace{3pt}+MD w/ Prompt Tune   & 8.23            & 1.67            & 9.21            & 3.70            & 5.70              \\
\hspace{3pt}+MD w/ Prefix Tune   & 6.50            & 3.33            & 4.94            & 3.56            & 4.58              \\ \hline
\end{tabular}
}
\label{table1}
\end{table}

\begin{figure}
	\centering
		\includegraphics[scale=.26]{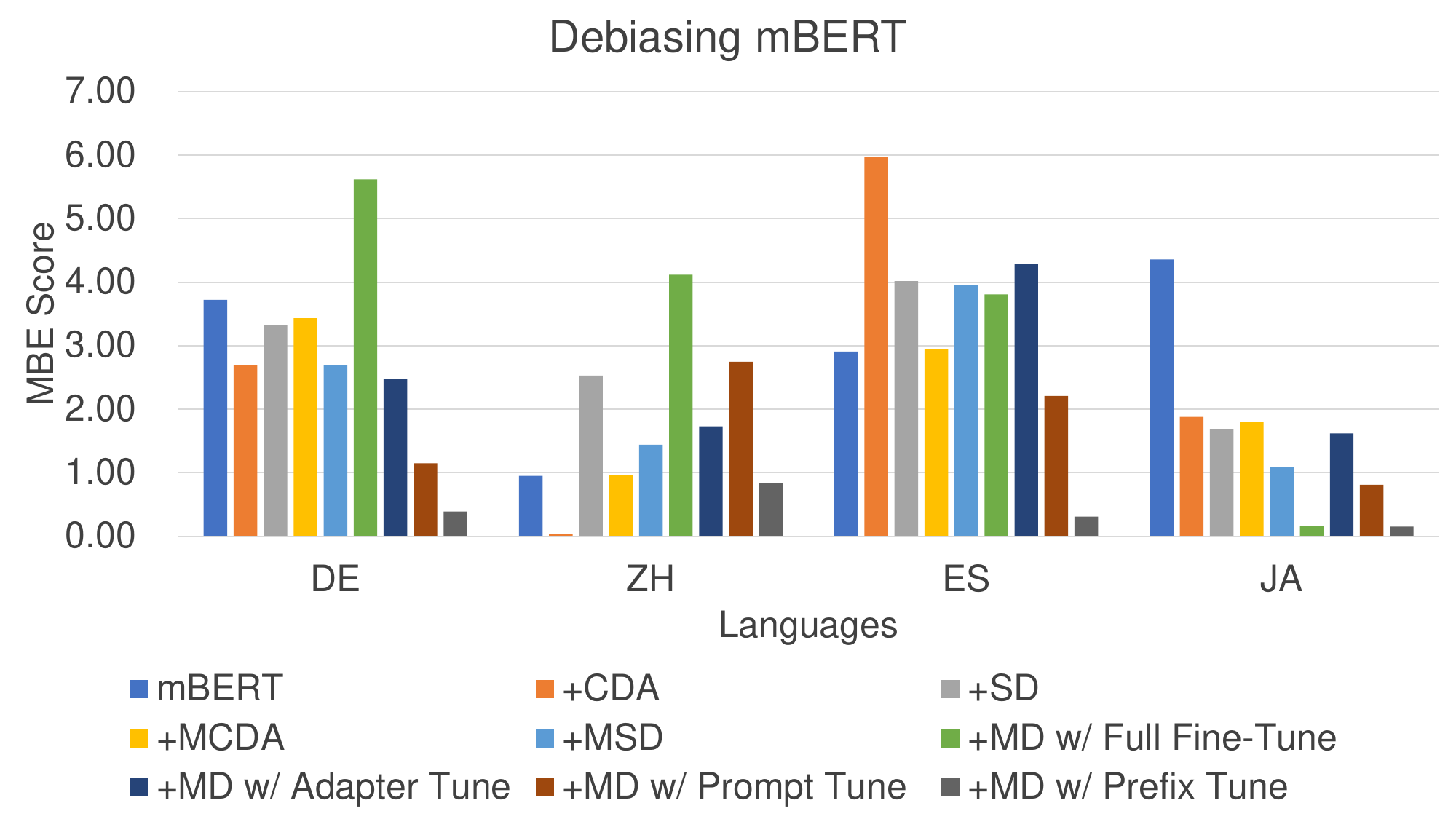}
	\caption{MBE bias scores of various debiasing methods on the mBERT model.}
	\label{FIG:2}
\end{figure}

\begin{figure}
	\centering
		\includegraphics[scale=.26]{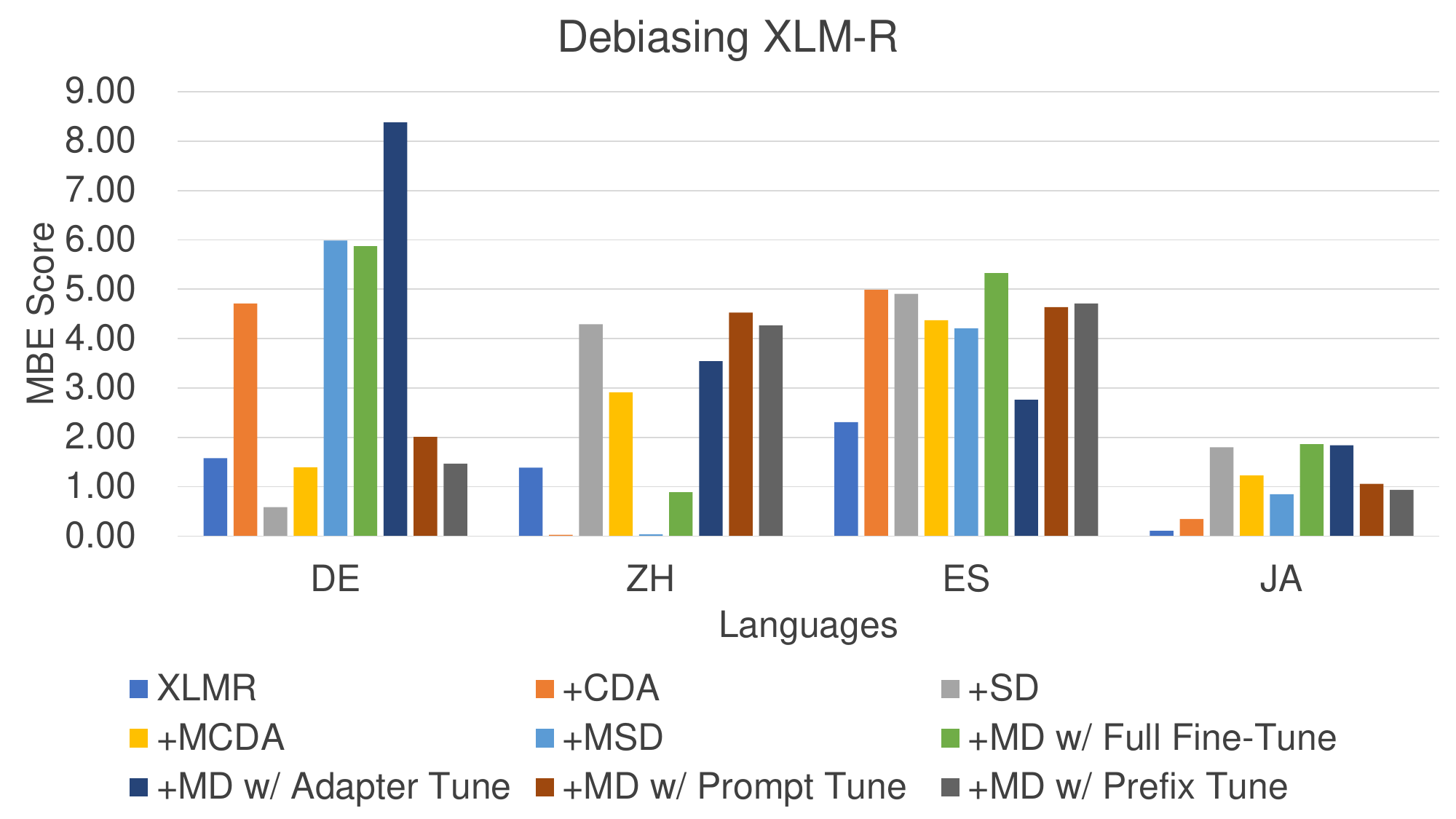}
	\caption{MBE bias scores of various debiasing methods on the XLM-R model.}
	\label{FIG:3}
\end{figure}

\textbf{The results from the MBE further confirm that multilingual debiasing methods are more effective at addressing gender-sensitive attributes.} As shown in Fig. \ref{FIG:2}, nearly all methods significantly reduced gender bias in mBERT for Chinese and Japanese, with multilingual approaches outperforming monolingual CDA and Self-Debias. In German and Spanish, monolingual CDA showed inconsistent results, and Self-Debias did not effectively mitigate bias. However, some multilingual debiasing methods such as MCDA and prefix-tuned MD consistently reduced gender bias in mBERT. Fig. \ref{FIG:3} demonstrates that the XLM-R model achieved satisfactory fairness levels without additional debiasing, making further improvements challenging. Independent use of CDA/MCDA or SD/MSD yielded unstable results in Chinese and German and performed poorly in Spanish and Japanese. While the MD method was not consistently effective across all languages, some combinations proved more stable than monolingual CDA and Self-Debias in specific languages. Consistent with Vashishtha et al.\cite{DBLP:conf/acl/VashishthaAS23}, our results suggest that multilingual debiasing, which integrates information across languages, holds promise for further exploration and refinement in future work.

\textbf{PEFT methods significantly enhance the debiasing capabilities of multilingual approaches.} Our analysis of the MBE evaluation results from two models reveals that the MD approach utilizing traditional full fine-tuning generally underperforms compared to the MD combinations that incorporate PEFT. In the mBERT tests, all languages except Japanese show a substantial improvement when using PEFT in the MD method compared to the full fine-tuning approach. A similar pattern is observed in the XLM-R tests, where PEFT methods outperform full fine-tuning in three out of four languages, with the exception of German. We attribute this success to the ability of PEFT methods to specifically modify the model's components related to sensitive attributes, allowing MCDA to be more effectively applied in the debiasing process.

\subsection{Results of Mitigating Racial and Religious Bias}
\textbf{In the evaluation of racial and religious biases, our multilingual debiasing methods also demonstrate superior effectiveness compared to the two monolingual methods.} As shown in Table \ref{table2}, both CDA and Self-Debias underperform against the baseline model when addressing racial sensitivity in mBERT. In contrast, the MD applied through full fine-tuning, prompt-tuning, and prefix-tuning, exhibit outstanding debiasing performance. This trend is also evident in the assessment of religious bias. Table \ref{table3} indicates that while the two monolingual debiasing methods achieve satisfactory results on the XLM-R model, our proposed multilingual approach significantly outperforms them.

\textbf{Self-Debias (SD) proves highly effective in reducing racial bias in the XLM-R model and religious bias in the mBERT model.} As illustrated in Table \ref{table2}, we note that nearly all methods show a decline in their effectiveness for debiasing racial attributes in Japanese, with SD being the notable exception. Similarly, Table \ref{table3} demonstrates that SD significantly outperforms other methods in mitigating religious bias in mBERT. These findings align with those of Meade et al. \cite{DBLP:conf/acl/MeadePR22}, which indicate that while CDA-based methods struggle with debiasing racial and religious attributes, SD retains a level of effectiveness. This is also why we consider incorporate MSD into our multilingual debiasing methods.

\begin{table}[]
\centering
\caption{Evaluation of Racial Bias on Multilingual CrowS-Pairs}
\resizebox{\linewidth}{!}{
\begin{tabular}{lccccc}
\hline
Model                  & DE     & ZH     & ES     & JA    & Avg. ($\downarrow$)  \\ \hline
mBERT                  & 4.74  & 15.00 & 5.00  & \textbf{3.33}  & 7.02  \\ \hline
\hspace{3pt}+CDA                   & \textbf{2.01}  & \textbf{1.67}  & 12.50 & 20.00 & 9.04  \\
\hspace{3pt}+SD                    & 7.24  & 12.50 & 4.17  & 5.83  & 7.44  \\ \hline
\hspace{3pt}+MCDA                  & 5.71  & 10.00 & 5.00  & 7.92  & 7.16  \\
\hspace{3pt}+MSD                   & 5.56  & 6.67  & 5.83  & 7.50  & 6.39  \\
\hspace{3pt}+MD w/ Full Fine-Tune & 2.93  & 6.67  & 3.33  & 8.33  & \textbf{5.32}  \\
\hspace{3pt}+MD w/ Adapter Tune  & 11.41 & 8.33  & 1.67  & 8.33  & 7.44  \\
\hspace{3pt}+MD w/ Prompt Tune   & 10.56 & 5.00  & 1.67  & 5.00  & 5.56  \\
\hspace{3pt}+MD w/ Prefix Tune   & 10.56 & 2.50  & \textbf{0.00}  & 9.17  & 5.56  \\ \hline
XLM-R                   & 16.37 & \textbf{0.00}  & 11.67 & 3.33  & 7.84  \\ \hline \hline
\hspace{3pt}+CDA                   & 15.56 & 12.50 & 9.17  & 5.83  & 10.76 \\
\hspace{3pt}+SD                    & 15.49 & 0.83  & \textbf{3.33}  & \textbf{1.67}  & \textbf{5.33}  \\ \hline
\hspace{3pt}+MCDA                  & 8.01  & 15.21 & 10.42 & 15.63 & 12.31 \\
\hspace{3pt}+MSD                   & 10.43 & 6.67  & 10.00 & 5.00  & 8.02  \\
\hspace{3pt}+MD w/ Full Fine-Tune & \textbf{0.53}  & 17.50 & 10.00 & 21.67 & 12.43 \\
\hspace{3pt}+MD w/ Adapter Tune  & 8.01  & 0.83  & \textbf{3.33}  & 17.50 & 7.42  \\
\hspace{3pt}+MD w/ Prompt Tune   & 7.88  & 1.67  & 4.17  & 10.00 & 5.93  \\
\hspace{3pt}+MD w/ Prefix Tune   & 0.56  & 13.33 & 14.17 & 12.50 & 10.14 \\ \hline
\end{tabular}
}
\label{table2}
\end{table}

\begin{table}[]
\centering
\caption{Evaluation of Religious Bias on Multilingual CrowS-Pairs}
\resizebox{\linewidth}{!}{
\begin{tabular}{lccccc}
\hline
Model                  & DE     & ZH    & ES    & JA    & Avg. ($\downarrow$) \\ \hline
mBERT                  & 19.17 & 18.65 & 22.33 & 4.19  & 16.08 \\ \hline
\hspace{3pt}+CDA                   & 22.50 & 22.07 & 23.21 & 9.25  & 19.26 \\
\hspace{3pt}+SD                    & \textbf{6.67}  & \textbf{6.77}  & \textbf{12.24} & \textbf{2.52}  & \textbf{7.05}  \\ \hline
\hspace{3pt}+MCDA                  & 17.71 & 16.57 & 22.55 & 3.36  & 15.05 \\
\hspace{3pt}+MSD                   & 15.83 & 8.46  & 15.53 & 3.37  & 10.80 \\
\hspace{3pt}+MD w/ Full Fine-Tune & 22.50 & 11.88 & 19.79 & 2.54  & 14.18 \\
\hspace{3pt}+MD w/ Adapter Tune  & 24.17 & 14.42 & 16.43 & 5.92  & 15.24 \\
\hspace{3pt}+MD w/ Prompt Tune   & 24.17 & 16.11 & 17.31 & 5.87  & 15.86 \\
\hspace{3pt}+MD w/ Prefix Tune   & 26.67 & 19.49 & 15.60 & 8.40  & 17.54 \\ \hline
XLM-R                   & 21.67 & 22.03 & 17.22 & \textbf{1.62}  & 15.64 \\ \hline \hline
\hspace{3pt}+CDA                   & 16.67 & 16.94 & 4.66  & 3.42  & 10.42 \\
\hspace{3pt}+SD                    & 16.67 & 20.36 & 6.30  & 7.63  & 12.74 \\ \hline
\hspace{3pt}+MCDA                  & 11.88 & 6.54  & 14.05 & 13.54 & 11.50 \\
\hspace{3pt}+MSD                  & 20.00 & 19.47 & 2.91  & 10.98 & 13.34 \\
\hspace{3pt}+MD w/ Full Fine-Tune & \textbf{4.17}  & 6.00  & \textbf{1.24}  & 9.34  & \textbf{5.19}  \\
\hspace{3pt}+MD w/ Adapter Tune  & 9.17  & \textbf{0.86}  & 10.51 & 10.19 & 7.68  \\
\hspace{3pt}+MD w/ Prompt Tune   & 13.33 & 10.13 & 7.03  & 8.48  & 9.74  \\
\hspace{3pt}+MD w/ Prefix Tune   & 7.50  & 1.67  & 7.09  & 7.65  & 5.98  \\ \hline
\end{tabular}
}
\label{table3}
\end{table}

\textbf{The Multiple-Debias method (MD), which combines MCDA and MSD, achieves the most effective debiasing results, with the debiasing effect of prompt-tuned MD being the most stable.} As shown in Table \ref{table2}, applying MCDA or MSD individually to mitigate racial bias in the XLM-R model does not surpass the baseline. However, their combination significantly improves performance, exceeding the baseline. Similarly, Table \ref{table3} shows that while MCDA or MSD alone offers some improvement in addressing religious bias, the combined approach achieves notably better results. Notably, only the method that integrates prompt-tuning with MD consistently outperforms the baseline across racial, religious, and gender debiasing.

\textbf{Integrating debiasing information from different languages can effectively improve the fairness of MPLMs.} We found that simply adding multilingual counterfactual data and multilingual prompts can achieve better results than monolingual debiasing methods. Similarly, Ahn et al. \cite{DBLP:conf/emnlp/AhnO21} observed that incorporating multilingual training data into mBERT acts as a form of debiasing. The multilingual counterfactual data we used for fine-tuning likely introduced valuable cross-lingual fairness signals, substantially improving the multilingual debiasing capabilities of our approach.

\section{Conclusion}
In this work, we addressed biases in multilingual settings by introducing Multiple-Debias, a full-process method for reducing gender, racial, and religious biases in multilingual pre-trained language models. Our approach combines MCDA and MSD across pre-processing and post-processing stages, with PEFT further enhancing effectiveness. We also expanded CrowS-Pairs into Multilingual CrowS-Pairs for German, Spanish, Chinese, and Japanese, enabling comprehensive bias evaluation across languages. A key finding is that debiasing with multilingual data achieves greater fairness than monolingual approaches alone, particularly when leveraging information across languages. We hope our work helps to guide future research in bias mitigation and inspires further advancements in multilingual fairness.

\section*{Acknowledgment}
% This work was supported by the National Natural Science Foundation of China (No. 62376062), the Ministry of Education of Humanities and Social Science Project (No. 23YJAZH220), the Philosophy and Social Sciences 14th Five-Year Plan Project of Guangdong Province (No. GD23CTS03), and the Guangdong Basic and Applied Basic Research Foundation of China (No. 2023A1515012718).

This work was supported by the National Natural Science Foundation of China (No. 62376062), the Ministry of Education of Humanities and Social Science Project (No. 23YJAZH220), and the Philosophy and Social Sciences 14th Five-Year Plan Project of Guangdong Province (No. GD23CTS03).

\bibliographystyle{IEEEtran}
\bibliography{main}

\end{document}